\documentclass[letterpaper, 10pt, conference]{ieeeconf}
\IEEEoverridecommandlockouts
\usepackage{graphicx}
\usepackage{amsmath}
\usepackage{amssymb}
\usepackage{color}
\usepackage{bbm}
\usepackage{algorithm}
\usepackage{algorithmic}
\usepackage{diagbox}
\usepackage{float}

\usepackage{subfigure}

\begin{document}
\title{
    Off-road Autonomous Vehicles Traversability Analysis and Trajectory Planning Based on Deep Inverse Reinforcement Learning 
}
\author{
    Zeyu~Zhu,
    Nan~Li,
    Ruoyu~Sun,
    Donghao~Xu,
    Huijing~Zhao%
\thanks{This work is supported by the National Natural Science Foundation of China under Grant(61973004). Z. Zhu, L. Nan, R. Sun, D. Xu and H. Zhao are with the Key Lab of Machine Perception (MOE), Peking University,
 Beijing, China.}
\thanks{Correspondence: H. Zhao, zhaohj@cis.pku.edu.cn.}
}

\maketitle
\thispagestyle{empty}
\pagestyle{empty}

\begin{abstract}
Terrain traversability analysis is a fundamental issue to achieve the autonomy of a robot at off-road environments. Geometry-based and appearance-based methods have been studied in decades, while behavior-based methods exploiting learning from demonstration (LfD) are new trends. Behavior-based methods learn cost functions that guide trajectory planning in compliance with experts' demonstrations, which can be more scalable to various scenes and driving behaviors. This research proposes a method of off-road traversability analysis and trajectory planning using Deep Maximum Entropy Inverse Reinforcement Learning. To incorporate the vehicle's kinematics while solving the problem of exponential increase of state-space complexity, two convolutional neural networks, i.e., RL ConvNet and Svf ConvNet, are developed to encode kinematics into convolution kernels and achieve efficient forward reinforcement learning. We conduct experiments in off-road environments. Scene maps are generated using 3D LiDAR data, and expert demonstrations are either the vehicle's real driving trajectories at the scene or synthesized ones to represent specific behaviors such as crossing negative obstacles. Different cost functions of traversability analysis are learned and tested at various scenes of capability in guiding the trajectory planning of different behaviors. We also demonstrate the peformance and computation efficiency of the proposed method. 
\end{abstract}
\section{Introduction and Related Works}
Terrain traversability analysis is a fundamental issue to achieve the autonomy of a robot at off-road environments. At such scenes, many methods for urban streets are not adaptive as there is no pavement or lane marking, no curb or other artificial objects to delimit road and no-road region, terrain surface is formed by natural objects that may have complex visual and geometric properties, etc. An extensive review of the challenges and literature works is given in \cite{Papadakis2013Terrain}. LiDARs and cameras have been used as the major sensors of online traversability analysis, where the mainstream methods are divided by \cite{Papadakis2013Terrain} into geometry-based and appearance-based ones.

\begin{figure}[h]
	\begin{center}
		\includegraphics[keepaspectratio=true,width=\linewidth]{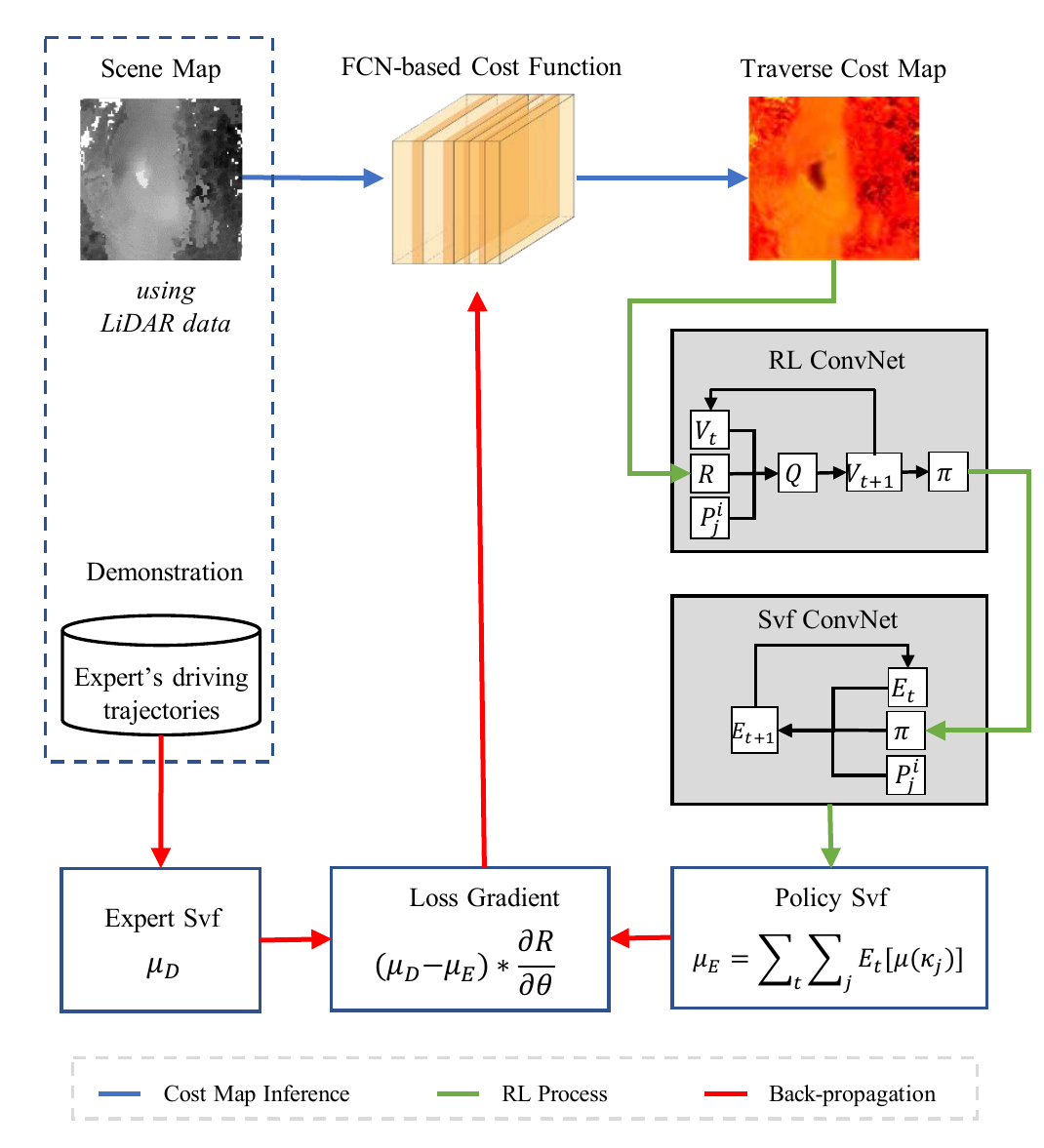}
	\end{center}
	\vspace{-5mm}
	\caption{Framework of our proposed method. The RL ConvNet and Svf ConvNet are our major contribution that incorporates vehicle kinematics and solves computation efficiency problem. Here “Svf” denotes state visiting frequency.}
	\label{IRL Training}
	\vspace{-6mm}
\end{figure}

Geometry-based methods generate a geometric representation of the world first using LiDAR or depth data, then assess traversability by comparing the geometric features such as height, roughness, slope, curvature and width with the vehicle's mechanical properties \cite{Lalonde2010Natural}\cite{Larson2011Lidar}\cite{6094685}.
 
Appearance-based methods assume that traversability is correlated with terrain appearance, and many learning-based approaches have been developed \cite{Labrosse2015Automatic}\cite{Mei2017Scene}. In order to improve far-field capability, methods are developed using underfoot or near field data to self-supervise the learning \cite{Howard2010Towards}\cite{Zhou2012Self}. Recently, deep neural networks are also employed to model the procedure \cite{7139749}\cite{Dan2017Find}\cite{Gao2019}, where in order to solve the problem of data annotation, semi-supervised learning methods are developed by incorporating the weakly supervised labels such as the vehicles' driving path.

Behavior-based method is a new trend in this field, which is inspired by the development of learning from demonstration (LfD) and promising results in recent years \cite{Argall2009A}.
Mainstream algorithms in LfD area can be approximately divided into two classes, Behavior Cloning (BC) \cite{Hayes1994A}\cite{Pomerleau}\cite{BojarskiTDFFGJM16} and Inverse Reinforcement Learning (IRL) \cite{Abbeel2004Apprenticeship}\cite{Ziebart2008}. Behavior Cloning directly learns a mapping from observation to action while IRL recovers the essential reward function behind expert demonstrations. Although earlier IRL algorithms use simple linear reward functions \cite{Abbeel2004Apprenticeship}\cite{Ziebart2008}\cite{Ratliff2006Maximum}, deep neural networks reward structures \cite{Wulfmeier2015Maximum}\cite{Finn2016Guided} are proposed later to model high-dimensional and non-linear process.
Compared with handcrafted cost and supervised-learning methods, IRL has better robustness and scalability \cite{Osa2018An}. Recently, deep maximum entropy IRL has been used to learn a traversable cost map for urban autonomous driving \cite{7759328}\cite{Wulfmeier2017Large}, and vehicle kinematics has also been considered in \cite{YanfuZhang2018} by converting history trajectory into new data channels, which are integrated with scene features to compose the input of a CNN based cost function. However in these works, vehicle kinematics is not incorporated in the forward reinforcement learning procedure, and the methods of value iteration and state visitation frequency estimation have poor efficiency. 

This research proposes a method of off-road traversability analysis and trajectory planning using Deep Maximum Entropy Inverse Reinforcement Learning. Novel contributions are that we encode vehicle kinematics into convolution kernels and propose two novel convolutional neural networks (RL ConvNet and Svf ConvNet) to achieve efficient forward reinforcement learning process, which solves the problem of exponential increase of state-space complexity. Experiments are conducted at off-road environments using real driving trajectories and synthesized ones that represent specific behaviors as demonstration. Results validate the performance and efficiency of our method.

This paper is organized as follows. The proposed methodology is described in Section.\ref{sec:methodology}. Experimental results are shown in Section.\ref{sec:experiment}. Finally, conclusion and future works are given in Section.~\ref{sec:conclusion and future works}.

\section{Methodology}
\label{sec:methodology}
As illustrated in Fig.~\ref{IRL Training}, this research proposes a deep inverse reinforcement learning framework for analyzing off-road autonomous vehicle traversability and planning trajectories, which incorporates kinematics and employs RL ConvNet and Svf ConvNet for efficient computation.

\subsection{Problem Formulation}
We formulate the process of autonomous vehicles navigating through off-road environment as Markov Decision Process(MDP), which can be defined as a tuple $(\mathcal{S},\mathcal{A},\mathcal{P},\gamma,\mathcal{R})$, where $\mathcal{S}$ denotes state space of the scene, $\mathcal{A}$ denotes action set of the autonomous vehicle, $\mathcal{P}$ denotes state transition probabilities, $\gamma \in [0,1)$ denotes discount factor and finally $\mathcal{R}$ denotes the traverse reward. Let $\cal{C}$ be traversability costs, ${\cal R}=-{\cal C}$, where the lower the costs, the higher the rewards.

Given demonstration samples set $D=\{ {\cal V}_i, \xi_i\}_{1:N_D}$, where at scene ${\cal V}_i$, the vehicle is driven through trajectory $\xi_i$ by a human expert.
A trajectory $\xi$ is a sequence of state-action pairs $\{(s_1, a_1), ..., (s_T, a_T)\}$, where actions $a_t$ are taken sequentially at states $s_t$. The reward value ${\cal R}(\xi)$ of a trajectory $\xi$ is simply the accumulative rewards (or negative costs) over all states that the trajectory traversed.
\begin{equation}
\label{eqn_reward_xi}
{\cal R}(\xi) = \displaystyle \sum_{t=1}^T \gamma^{t-1} {\cal R}(s_t) = - \sum_{t=1}^T \gamma^{t-1} {\cal C}(s_t)
\end{equation}

Let $f_\theta$ be a function to evaluate traversability cost ${\cal C}$ of a certain scene ${\cal V}$ with features $\phi ({\cal V})$, ${\cal C} = - {\cal R} = f_\theta (\phi ({\cal V}))$.
Following Wulfmeier et al. \cite{7759328}, we use grid maps to represent $\phi ({\cal S})$, ${\cal C}$ and ${\cal R}$, and a fully convolutional neural network(FCN) for $f_\theta$ with a parameter set $\theta$. 
It is assumed that human expert trajectories are intending to maximize rewards gain or minimizing traversability costs. Our goal is to learn a parameter set $\theta$ for $f_\theta$ from expert's demonstrations, so as to guide an autonomous agent to plan trajectories in similar ways as human drivers.

\subsection{Maximum Entropy Deep IRL}
Under the maximum entropy assumption, probability of a trajectory $\xi$ is estimated below, where trajectories with higher reward values are exponentially more preferrable \cite{Ziebart2008}. $Z$ is an integral term which is usually referred to as partition function.

\begin{eqnarray}
\label{eqn_prob_xi}
p(\xi|{\cal V};\theta) &=& \displaystyle \frac{\exp({\cal R}(\xi|{\cal V};\theta))}{Z({\cal V};\theta)} \\
Z({\cal V};\theta) &=& \int \exp {\cal R}(\xi|{\cal V};\theta)d\xi \nonumber
\end{eqnarray}

Given demonstration samples set $D$, learning $\theta$ can be formulated as maximizing the following log-likelihood problem.

\begin{eqnarray}
\label{eqn_optimal_theta}
L(\theta) &=& \displaystyle \prod_{i=1}^{N_D}p(\xi_i^D|{\cal V}_i^D;\theta)\\
\theta^* &=&\displaystyle \arg\max_\theta L(\theta) \nonumber \\
&=& \displaystyle \arg\max_\theta \sum_{i=1}^{N_D} \log p(\xi_i^D|{\cal V}_i^D;\theta)
\end{eqnarray}

Let $\mu_D$ and $\mu_E$ denote the state visiting frequencies of human expert drivers' policy and optimal policy recovered from reward function respectively, where $\mu_D$ is approximated from human demonstration samples, while $\mu_E$ is estimated by solving the MDP. According to Ziebart et al. \cite{Ziebart2008} and Wulfmeier et al. \cite{7759328}, optimizing $\theta$ is conducted by back-propagating the following loss gradient.
\begin{equation}
\label{eqn_gradient}
\bigtriangledown_\theta L(\theta) = (\mu_D - \mu_E) \bigtriangledown_\theta \mathcal{R} = (\mu_E - \mu_D) \bigtriangledown_\theta f_\theta
\end{equation} 

Hence, given the current parameter set $\theta$, the following steps are taken at each iteration for optimization. The processing flow is as follows.

\begin{enumerate}
	\item Estimating traversability cost ${\cal C} = f_\theta (\phi ({\cal S}))$, and let ${\cal R} = -{\cal C}$;
	\item Reinforcement learning to find an optimal policy $\pi$ (Algorithm 1)
	\item Computing expected state visitation frequency $\mu_E$ (Algorithm 2)
	\item Computing expert's state visitation frequency $\mu_D$ from demonstrated trajectories
	%\item Updating $\theta$ with loss gradient given by Eqn.~\ref{eqn_gradient}
	\item Optimizing $\theta' = \theta - \lambda \bigtriangledown_{\theta} L(\theta)$ by Eqn. \ref{eqn_gradient}, where $\lambda$ is a learning rate.
\end{enumerate}
\begin{algorithm}
\label{alg_1}
\caption{Value iteration (Original method)}
\begin{algorithmic}[1]
	\REQUIRE reward ${\cal R}$, transition probability ${\cal P}$
	\ENSURE optimal stochastic policy $\pi$
	\STATE Initialize: $V_0(s)=-\infty, V_0(s_{goal})=0$
	\FOR {$t=1:K$}
	\FOR {$s \in \mathcal{S}$}
	\FOR {$a \in \mathcal{A}$}
	\STATE $Q_t(s,a) = \displaystyle {\cal R}(s) + \gamma \sum_{s'} {\cal P}(s'|s,a)*V_{t-1}(s')$
	\ENDFOR
	\STATE $V_t(s) = \displaystyle Softmax_{a} \ Q_t(s,a)$ 
	\ENDFOR
	\ENDFOR 
	\STATE $\pi(a|s) = exp (Q_K(s,a)-V_K(s))$
\end{algorithmic}
\end{algorithm}

\begin{algorithm}
\label{alg_2}
\caption{Computing expected state visitation frequency (Original method)}
\begin{algorithmic}[1]
	\REQUIRE stochastic policy $\pi$, transition probability ${\cal P}$, initial state distribution $D(s)$
	\ENSURE expected state visiting frequency $\mu_E$
	\STATE Initialize: $\mathbf{E}_1[\mu(s)] = \displaystyle D(s)$
	\FOR { $t=1:T$ }
	\FOR {$s \in \mathcal{S}$}	
	\STATE $\mathbf{E}_{t+1}[\mu(s)]=\displaystyle \sum_{s',a} {\cal P}(s'|s,a) \pi(a|s') \mathbf{E}_t[\mu(s')]$
	\ENDFOR
	\ENDFOR
	\STATE $\mu_E = \mathbf{E}[\mu(s)] = \sum_t \mathbf{E}_{t}[\mu(s)]$
\end{algorithmic}
\end{algorithm}
\begin{figure*}[ht]
	\centering

	\subfigure[RL ConvNet]{
		\label{RLConvNet}
		\begin{minipage}[t]{0.50\linewidth}
			\centering
			\includegraphics[width=\linewidth,keepaspectratio=true]{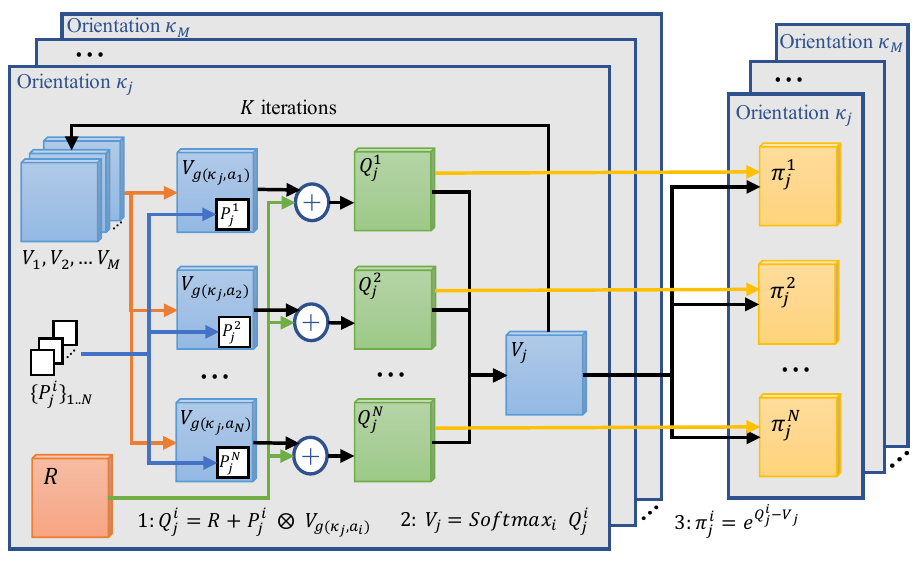}
			%\caption{RL ConvNet}
		\end{minipage}%
	}%
	\hspace{5mm}
	\subfigure[Svf ConvNet]{
		\label{SvfConvNet}
		\begin{minipage}[t]{0.45\linewidth}
			\centering
			\includegraphics[width=\linewidth,keepaspectratio=true]{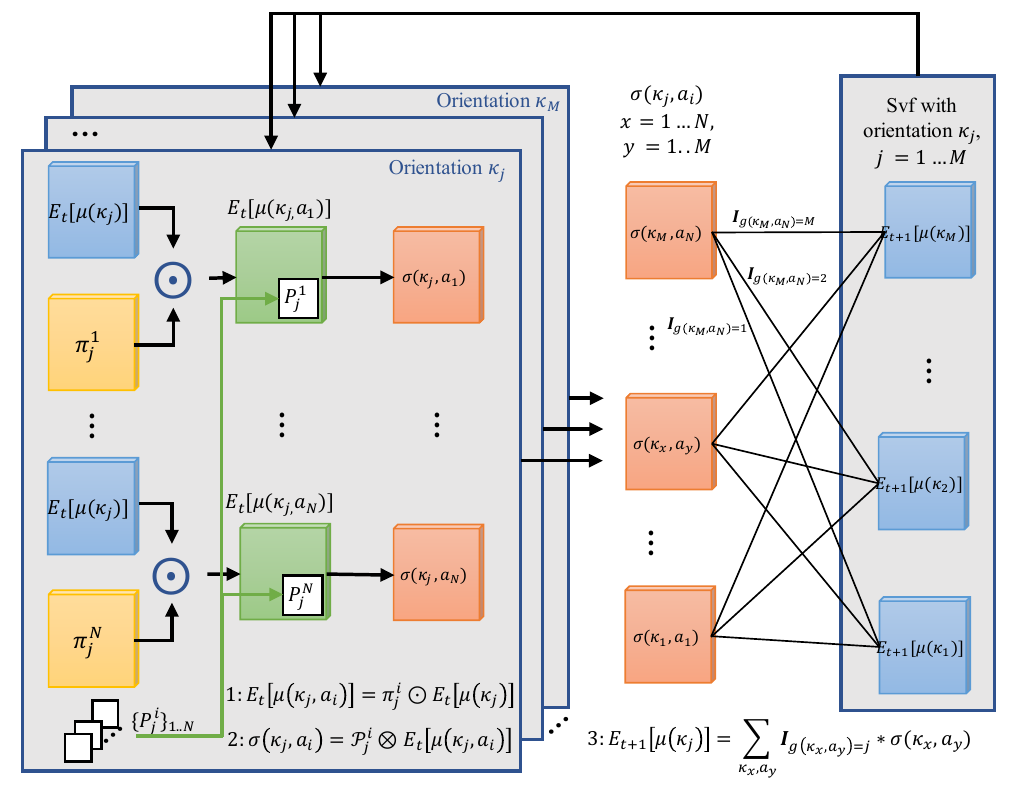}
			%\caption{fig2}
		\end{minipage}%
	}%
	\centering
	\vspace{-2mm}
	\caption{(a):Structure of our proposed RL ConvNet. (b):Structure of our proposed Svf ConvNet. $R$ denotes reward map(negative to cost). $V_j$ denotes value map with orientation $\kappa_j$. $\{P_j^i\}_{1..N}$ denotes transition kernels corresponding to actions $a_i,i=1..N.$ $Q_j^i$ and $\pi_j^i$ denotes Q maps and stochastic policy maps corresponding to orientation $\kappa_j$ and action $a_i$ respectively. $E[\mu(\kappa_j)]$ and $E[\mu(\kappa_j,a_i)]$ denotes state visiting frequency for orientation $\kappa_j$ and orientation-action pair $(\kappa_j,a_i)$ respectively.}
	\vspace{-6mm}
\end{figure*}

\subsection{Incorporating Vehicle Kinematics}
However, the following problems remain. First, vehicle kinematics is non-holonomic. Incorporating non-holonomic constraints is vital to plan trajectories which are physically operational by vehicles. Second, traditional value iteration and state visiting frequency estimation are time-consuming. Especially, incorporating kinematics comes with more state dimensions, resulting in an exponential increase in computation complexity.

Tamar et al. \cite{NIPS2016_6046} proposed a convolutional network structure for value iteration process, where previous value $V_{i-1}$ and reward ${\cal R}$ are passed through a convolution layer and max-pooling layer, each channel in the convolution output represents the $Q$ function of a specific action, and convolution kernel weights corresponds to the discounted transition probabilities. Thus by recurrently applying a convolution layer $K$ times, value iteration is efficiently performed with significant reduction of computation costs.

Inspired by the idea, we propose RL ConvNet and Svf ConvNet for both incorporating vehicle kinematics and achieving efficient computation at the same time.  
\subsubsection{RL ConvNet}
~\

We consider modeling kinematic constraints of vehicles' orientation. Let ${\cal A} = \{a_1,...,a_N\}$ and ${\cal K} = \{\kappa_1,...,\kappa_M \}$ be set of vehicles' discrete actions and orientations respectively. We assume that the vehicles' orientation constrains vehicles' state transition probability under a certain action, hence a set of convolutional kernels $\{{\cal P}_j^i \}_{N \times M}$ is defined, where ${\cal P}_j^i$  is a kernel with the weights corresponding to the discounted transition probabilities after taking action $a_i$ under orientation $\kappa_j$. To handle the exponential increase of computation complexity coming with incorporating kinematics in states without explicit performance degradation, we constrain the transition probability of vehicle orientations to be deterministic. More specific, we define the function $g(\kappa,a): \cal K \times \cal A \mapsto$ $\{1,2,...,M\}$, where $g(\kappa,a)$ represents the next step vehicle orientation index after taking action $a$ under current orientation $\kappa$. For convolution purpose, grid maps are employed to represent $V, Q,\pi$ for value, $Q$ and policy functions, and a grid pixel corresponds to a 2D location at the scene. Let $V_{j}$ be a value map for vehicle orientation $\kappa_j$, where for any pixel $s$, $V_{j}(s)$ is the expected maximal value  starting from the location $s$ with orientation $\kappa_j$. $Q$ and $\pi$ are defined in similar ways, which are related to both vehicle orientation and action. Let $Q_j^i$ and $\pi_j^i$ be a $Q$ and a policy map of vehicle orientation $\kappa_j$ and action $a_i$, where for any pixel $s$, $Q_j^i(s)$ is the expected long-term reward and $\pi_j^i(s)$ is the stochastic policy probability if action $a_i$ is taken when the vehicle is at the location $s$ and orientation $\kappa_j$.
\begin{algorithm}
	\label{alg_3}
	
	\caption{RL ConvNet}
	%Convolutional value iteration incorporating vehicle kinematics}
	\begin{algorithmic}[1]
		\REQUIRE reward ${\cal R}$, transition probability $\{{\cal P}_j^i\}_{N \times M}$, function $g(\kappa, a)$
		\ENSURE optimal stochastic policy $\pi_j^i$
		\STATE Initialize: $V_j = -\infty$
		\FOR {$t=1:K$}
		\FOR {$j=1:M$}
		\STATE $V_j(s_{goal})=0$
		\ENDFOR
		\FOR {$j=1:M$}
		\FOR {$i=1:N$}
		\STATE $Q_j^i = \displaystyle {\cal R} + {\cal P}_j^i \otimes V_{g(\kappa_j,a_i)}$ 	  
		\ENDFOR
		\STATE $V_j = \displaystyle Softmax_i \ Q_j^i$
		\ENDFOR
		\ENDFOR
		\STATE $\pi_j^i = \exp(Q_j^i - V_j)$
	\end{algorithmic}
	
\end{algorithm}
 Hence the original value iteration algorithm can be converted to Algorithm 3, which incorporates vehicle kinematics and accelerates computation through convolution. 

As is illustrated in Fig.~\ref{RLConvNet}, for a particular orientation $\kappa_j$, each $V_{g(\kappa_j,a_i)}, i=1,..., N$ is passed first through a convolution layer with kernel ${\cal P}_j^i$ corresponding to orientation $\kappa_j$ and action $a_i$, then through a pixel-wise addition layer with ${\cal R}$ to estimate $Q_j^i$.
% $V_{g(\kappa_j,a_i)}$ are passed through each ${\cal P}_j^i$ of orientation $\kappa_j$, hence a set of $\{Q_j^i|i = 1,...,N\}$ is obtained. 
 All ${Q_j^i}$s of orientation $\kappa_j$ are then passed through a Softmax layer to obtain updated $V_j$. These operations iterate $K$ times until $V_j$ converges. Since then, a sequence of policy maps $\{\pi_j^i\}, i = 1,..., N$ are estimated corresponding to the particular orientation $\kappa_j$ and each of the actions $a_i$, which are matrix operations. These operations are conducted for the $V_j$ of each a particular orientation $\kappa_j, j = 1,..., M$, hence a set of policy maps $\{\pi_j^i\}_{N \times M}$ are obtained corresponding to each pair of $\kappa_j$ and $a_i$.                                             

\subsubsection{Svf ConvNet}
~\

State visiting frequency is also represented as a grid map. Let $\mathbf{E}_t[\mu(\kappa_j,a_i)]$ be orientation-action state visiting frequency, where each pixel value is the expected frequency that action $a_i$ is taken under current orientation $\kappa_j$ at the corresponding 2D location. Similarly, $\mathbf{E}_t[\mu(\kappa_j)]$ denotes the expected state visitation frequency for vehicle orientation $\kappa_j$, which is simply sum of orientation-action state visiting frequency. Denote $\odot$ as pixel-wise multiplication of two grid maps, $\mathbf{E}_t[\mu(\kappa_j,a_i)]$ is calculated as follows: 
\begin{equation}
\mathbf{E}_{t}[\mu(\kappa_j,a_i)] =  \pi_j^i \odot \mathbf{E}_{t}[\mu(\kappa_j)]
\end{equation}

Hence the original algorithm of computing expected state visitation frequency can be converted to Algorithm 4, which incorporates vehicle kinematics and computes efficiently through convolution.

\begin{algorithm}
	\label{alg_4}
	\caption{Svf ConvNet}
		%Convolutional state visitation frequency computation incorporating vehicle kinematics}
	\begin{algorithmic}[1]
		\REQUIRE stochastic policy $\pi_j^i$, transition probability ${\cal P}_j^i$, initial state distribution $D(\kappa)$, function $g(\kappa, a)$
		\ENSURE expected state visiting frequency $\mu_E$
		\STATE Initialize: $\mathbf{E}_1[\mu(\kappa)] = \displaystyle D(\kappa)$
		\FOR { $t=1:T$ }
			\FOR {$j=1:M$}
				\FOR {$i=1:N$}
					\STATE $\mathbf{E}_{t}[\mu(\kappa_j,a_i)] =  \displaystyle \pi_j^i \odot \mathbf{E}_{t}[\mu(\kappa_j)]$ 
					\STATE $\sigma(\kappa_j,a_i) = \displaystyle {\cal P}^i_j \otimes \mathbf{E}_{t}[\mu(\kappa_j,a_i)]$
				\ENDFOR
			\ENDFOR
			\FOR {$j=1:M$}
				\STATE $\mathbf{E}_{t+1}[\mu(\kappa_j)] = \displaystyle \sum_{\kappa_x,a_y}\mathbf{I}_{g(\kappa_x,a_y)=j} * \sigma(\kappa_x,a_y)$ \\ %\COMMENT {$\mathbf{I}$ denotes indicator function}
			\ENDFOR
		\ENDFOR
		\STATE $\mu_E = \sum_t \sum_j \mathbf{E}_{t}[\mu(k_j)]$
	\end{algorithmic}
\end{algorithm}

Fig.~\ref{SvfConvNet} illustrats the workflow. For any particular orientation $\kappa_j$, policy $\pi_j^i$ of action $a_i$ and orientation $k_j$ is passed through a multiplication layer with $\mathbf{E}_t[\mu(k_j)]$. The resultant orientation-action state visiting frequency grid map is then passed through a convolution layer with
kernel ${\cal P}_j^i$ to obtain $\sigma(\kappa_j,a_i)$.
All $\sigma(\kappa_x,a_y)$s, $x=1,..,M$ and $y=1,..,N$ are passed through a weighted multiplication layer to find updated orientation state visiting frequency $\mathbf{E}_{t+1}[\mu(\kappa_j)]$s, where the weights are $\mathbf{I}_{g(\kappa_x,a_y)=j}$.

\section{Experiment}
\label{sec:experiment}
\subsection{Experimental Design and Dataset}
\begin{table*}[h]
	\caption{\label{Tab_Exp_design}Experimental Design}
	\begin{center}
		\begin{tabular}{c|c|c|c}
			\hline
			&  Scene & Expert's Behavior & Learnt Cost Function \\
			\hline
			\hline
			Exp.1 (E1)&straight and flat road scenes& normal driving behavior &$\mathbf{F_{onroad}}$\\
			\hline
			Exp.2 (E2)&negative obstacles scenes& avoid the negative obstacles &$\mathbf{F_{avoid-obs}}$\\
			\hline
			Exp.3 (E3)&negative obstacles scenes& cross negative obstacles if they block the way &$\mathbf{F_{cross-obs1}}$\\
			\hline
			Exp.4 (E4)&negative obstacles scenes& cross all negative obstacles on the road &$\mathbf{F_{cross-obs2}}$\\
			\hline
		\end{tabular}
	\end{center}
\vspace{-5mm}
\end{table*}
As shown in Tab.~\ref{Tab_Exp_design}, we design four experiments to examine the proposed method's learning capability of different experts' driving behaviors.
In Exp.1, an expert demonstrates normal driving behaviors on straight and flat roads and a cost function $\mathbf{F_{onroad}}$ is obtained as the result of learning, which is used as a baseline of other experiments. Exp.2-4 are conducted at negative obstacle scenes, where the expert's behavior is to avoid the negative obstacles in Exp.2, cross negative obstacles only if they block the way in Exp.3, while cross all negative obstacles on the road in Exp.4.
\begin{figure}[h]
	\centering
	\begin{center}
		\includegraphics[width=\linewidth,keepaspectratio=true]{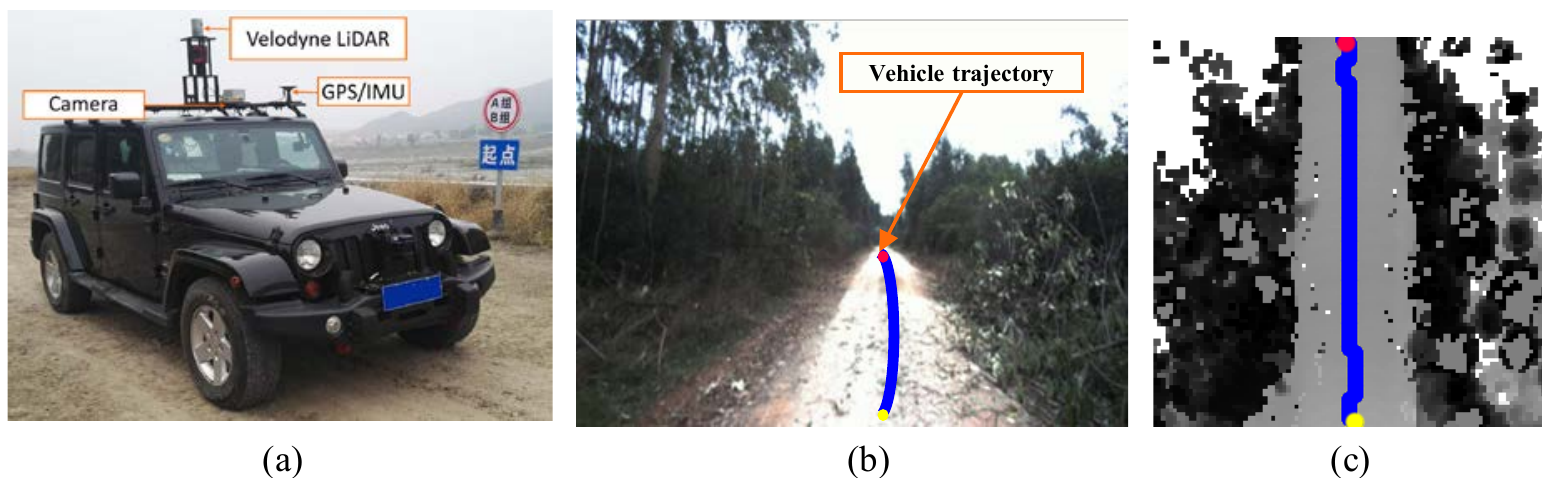}
	\end{center}
	\vspace{-5mm}
	\caption{ Our experimental vehicle platform and scene map example (a)Data collection platform(POSS-V) (b) Example RGB view of off-road environment with expert trajectory projected onto it (c) 2D scene grid map}
	\label{dataset_example}
	\vspace{-5mm}
\end{figure}

Data collection is conducted at off-road environments using an instrumented vehicle shown in Fig.\ref{dataset_example}(a).
The vehicle has a Velodyne HDL-64 LiDAR to map scene features, a GPS/IMU system to capture expert's driving trajectories, and a front-view monocular camera for visualization only.

We generate a dataset where each frame has a scene map, a demonstration trajectory crossing the scene, a start and a goal point which are defined by the trajectory points that enter and leave the scene. The scene map is a 100 $\times$ 100 2D grid one with a resolution of 0.25 meters. It is generated using LiDAR data and each grid is assigned the height value of the highest LiDAR points projected to the cell. The whole dataset has a total of 2388 scene maps, and the expert's real driving trajectories are used as the demonstrations for Exp.1 and Exp.2. Assume that we have a vehicle of stronger mobility and the expert chooses to cross all or some of the negative obstacles for efficiency. We use the real scene maps of negative obstacles, but synthesize demonstration trajectories in compliance with the defined behaviors of Exp.3 and Exp.4.

There are 320 frames of straight and flat road scenes in Exp.1 and trained a cost function $\mathbf{F_{onroad}}$. There are 320 frames of negative obstacle scenes but trajectories demonstrating different behaviors in Exp.2, 3 and 4, and trained cost functions $\mathbf{F_{avoid-obs}}$, $\mathbf{F_{cross-obs1}}$, $\mathbf{F_{cross-obs2}}$ respectively. The rest scenes are used for testing, where the focus is the comparison of different cost functions and the planned trajectories at the same scene.

\subsection{Implementation Details}
\begin{figure}[h]
	\begin{center}
		\includegraphics[keepaspectratio=true,width=0.9\linewidth]{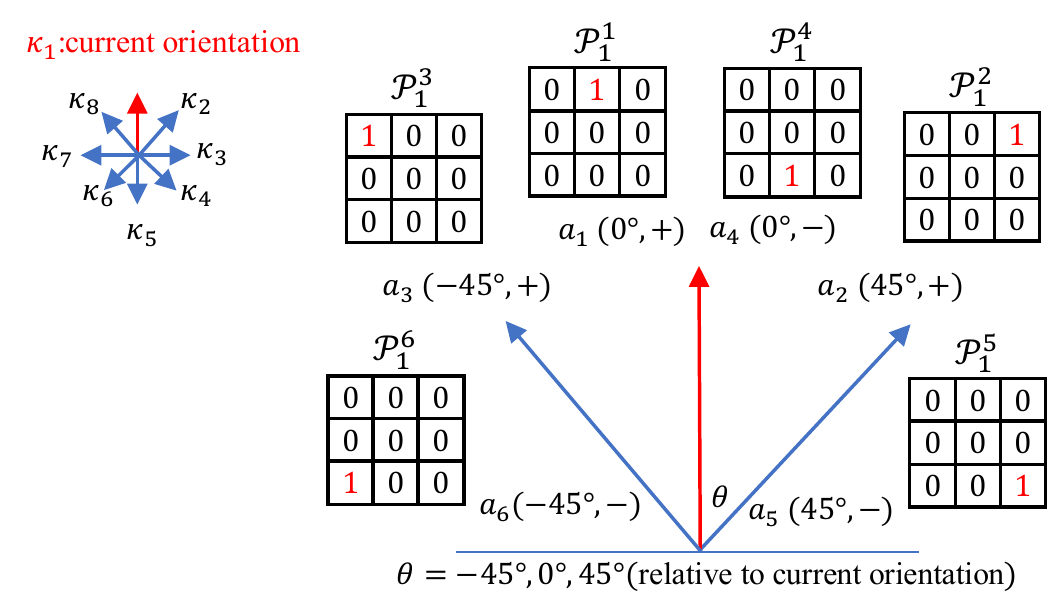}
	\end{center}
	\vspace{-5mm}
	\caption{Illustration of part of transition convolutional kernels used in experiment. "+/-" represents forward/backward.}
	
	\label{transition}
	\vspace{-2mm}
\end{figure}
\subsubsection{State, Action Space and Transition kernels}
~\

In our implementation, state is the current 2D position of the vehicle in the scene map. We use discrete actions and orientations shown in Fig.~\ref{transition}. The orientation is discretized into eight directions with 45 degrees as interval to cover the full range. The actions are simplified to be combinations of steering angle(-45, 0, 45 degrees relative to current orientation) and driving forward or backward. The transition kernels corresponding to orientation $\kappa_1$ are illustrated in Fig.~\ref{transition}.  
%More details about function $g(\kappa,a): \cal K \times \cal A \mapsto$ $\{1,2,...,M\}$ can be found in Appendix.
\subsubsection{Network and Training Configuration}
~\

In the experiment, we adopt a simple five-layer fully convolutional network (FCN) structure which takes processed lidar feature map as input. We use size $5\times5$ and $3\times3$ for convolution kernels in FCN. As for RL ConvNet and Svf ConvNet, we set the number of value iterations $ K$ to 150 and number of svf iterations $T$ to 120, which are experimentally observed to execute effective reinforcement learning process and svf computation. The network is trained with the Adam optimizer with initial learning rate 1e-4 and learning rate decay 0.99. The batch training is also employed with batch size 5, which proved in practice more stable than updating weights based on a single demonstration.
\begin{figure}[h]
	\centering
	\begin{center}
		\includegraphics[width=\linewidth,keepaspectratio= true]{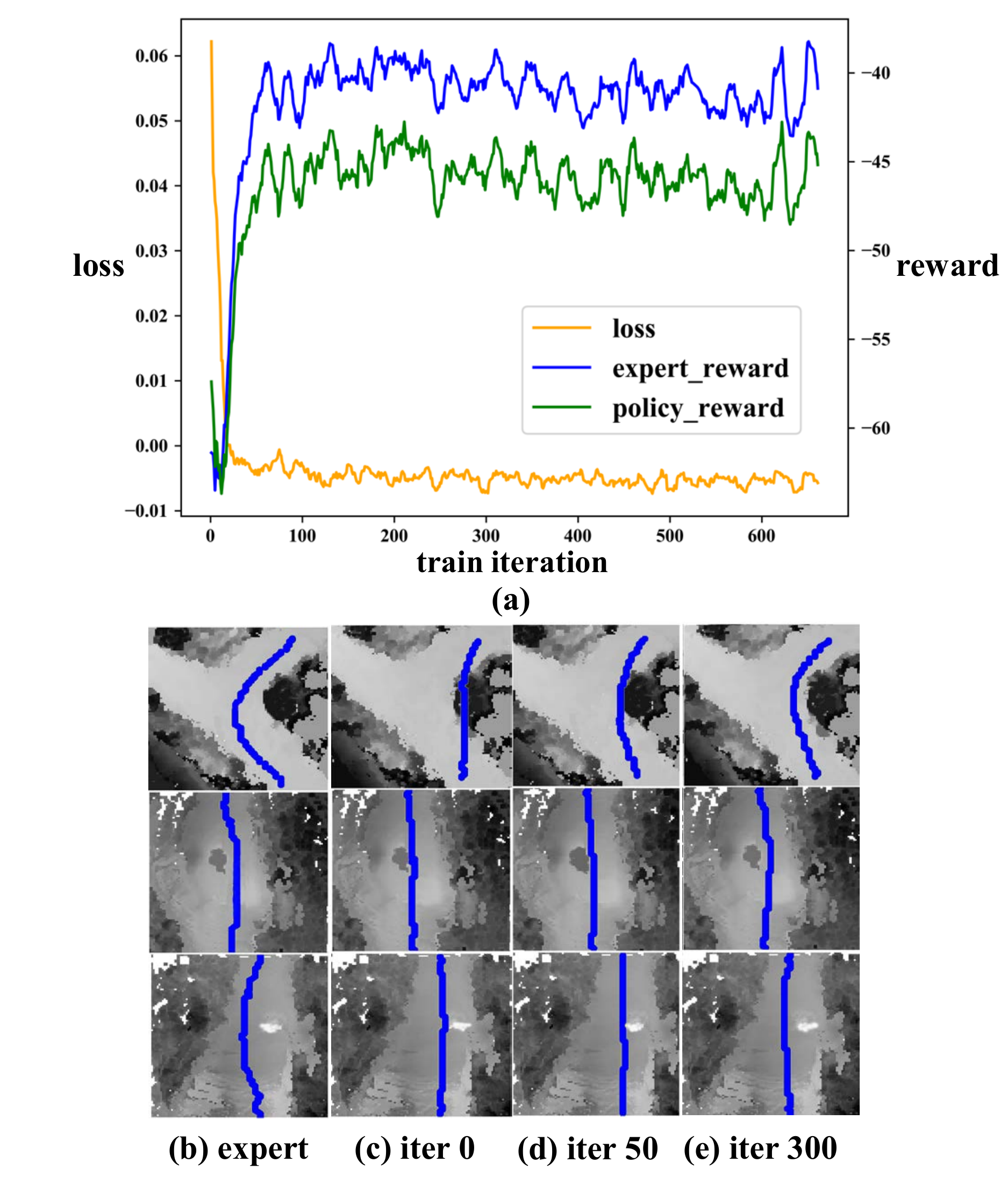}
	\end{center}
	\vspace{-5mm}
	\caption{Training process visualization. The loss curve, reward of expert's and policy's trajectories are shown in (a). Samples of expert's demonstration trajectories are shown in (b) while policy generated trajectories at different training stages are shown in (c)-(e).}
	\label{train_visual}
	\vspace{-6mm}
\end{figure}
\subsection{Training Results}
The training process visualization is shown in Fig.~\ref{train_visual}. During the training process, we periodically evaluate the expert's trajectory reward and policy's trajectory reward using Eqn.~\ref{eqn_reward_xi}. Average discounted cumulative reward of 30 trajectories randomly sampled from learned policy by simulation is used. As is shown in Fig.~\ref{train_visual}(a), the loss value decreases and tend to converge after 200 iterations. Expert's and policy's rewards are getting higher while policy reward is a little lower than expert's. Samples of expert's demonstration trajectories are shown in Fig.~\ref{train_visual}(b) while policy generated trajectories at different training stages are shown in Fig.~\ref{train_visual}(c)-(e). One can explicitly find that as iterations increase, similarity between expert's and policy's trajectories is becoming higher, which validates that learned reward guides trajectory planning successfully in compliance with human's behavior.
\begin{figure}[h]
	\centering
	\includegraphics[keepaspectratio=true,width = \linewidth]{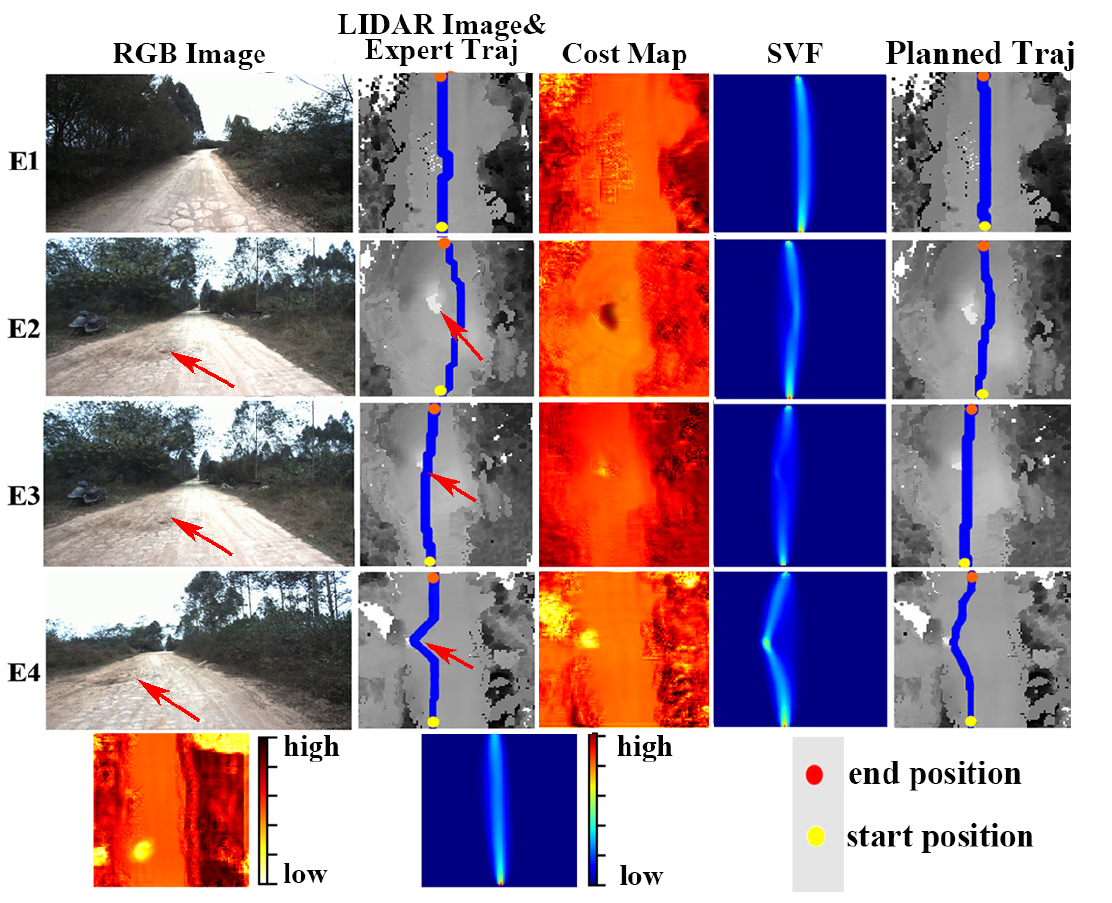}
	\vspace{-8mm}
	\caption{Training result visualization of four experiments(most importantly, experiments here are not to find the best cost map, but to demonstrate the learning capability of different behavior). Columns from left to right: RGB image for reference. LiDAR image with demonstration trajectory projected onto it. Learned cost map after training. State visiting frequency (Svf) map. Planned trajectory(in blue). As for Exp.2-4, negative obstacles can be found on the LiDAR map (\textbf{see the red arrows}). The learned cost maps have different traverse cost evaluation of these holes.}
	\label{training_result}
	\vspace{-2mm}
\end{figure}

\begin{figure*}[h]
	
	\centering
	\includegraphics[width=0.95\linewidth,keepaspectratio=true]{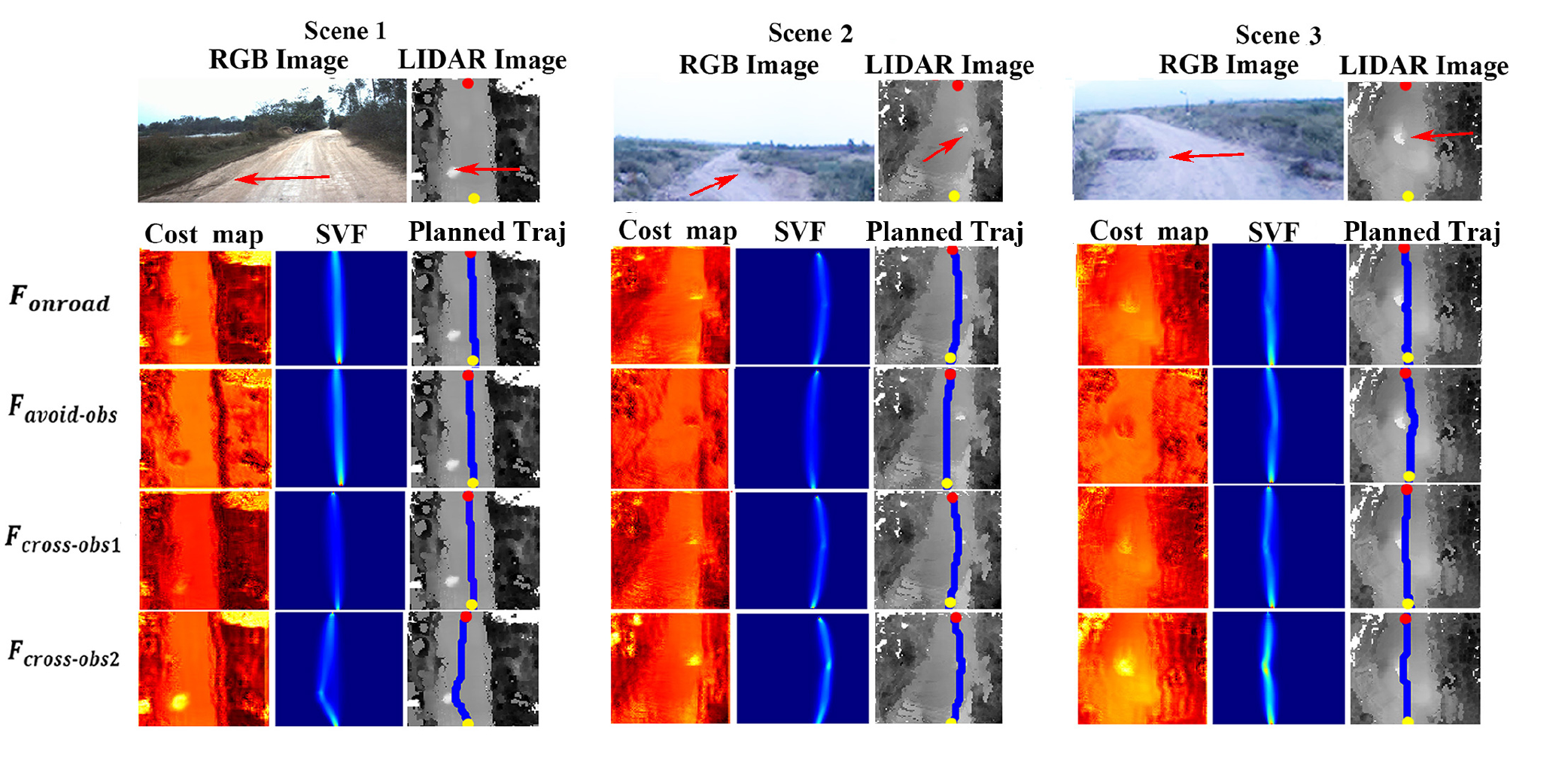}
	\vspace{-8mm}	
	\caption{Testing result visualization of different scenes. Due to limited page space, annotations can be found in Fig.~\ref{training_result}, and are omitted here.}
	
	\label{testing_result}
	\vspace{-4mm}
\end{figure*}
More visual evaluation results of our four experiments can be found in Fig.~\ref{training_result}. Four learned cost maps all successfully capture the high traverse cost feature of positive obstacles(i.e., trees or bushes) but differ in the assessment of negative obstacles(pits on the road). Demonstrated by human drivers' avoid-negative-obstacle trajectories, Exp.2 learns that negative obstacles have higher cost compared with flat roads. Exp.3 and Exp.4 have opposite results given cross-negative-obstacle trajectories as demonstration. Negative obstacles in cost maps of Exp.3 and Exp.4 have relatively lower costs compared to flat roads, resulting in cross holes behavior of planned trajectories. One interesting fact is that the result of Exp.4 assigns lower costs to holes than Exp.3, showing more preference for negative obstacles.

\vspace{-2mm}
\subsection{Testing Results}
Visualization of testing results on different scenes are shown in Fig.~\ref{testing_result}. We demonstrate the learning capability and scalability by comparing the behavior of trajectories generated by different learned cost functions. As shown in Fig.~\ref{testing_result}, $\mathbf{F_{onroad}}$ fail to handle scenes where negative obstacles exist and planned trajectories cross them. On opposite, by demonstrating avoiding obstacles behavior in Exp.2, $\mathbf{F_{avoid-obs}}$ shows good performance under these scenes and successfully avoids negative obstacles. By comparison, $\mathbf{F_{cross-obs1}}$ and $\mathbf{F_{cross-obs2}}$ show strong learning capability as they produce cost maps that assign low traverse cost to negative obstacles. Compared to $\mathbf{F_{cross-obs1}}$, $\mathbf{F_{cross-obs2}}$ assigns much lower cost to negative obstacles since it is demonstrated by trajectories more preferable for them. The results show that our method has a strong learning capability of different behaviors and is scalable to different scenes. 

	To evaluate the accuracy of replicating human behavior, we use the Hausdorff Distance (HD) \cite{Kitani2012Activity} as our metric. The HD metric represents a spatial similarity between expert
demonstrations and trajectories sampled with the learned policy. We use the average HD and 30 trajectories randomly sampled from learned policy are used for each single test. We choose deep maximum entropy IRL proposed by Wulfmeier et al. \cite{7759328} as baseline which presents good performance in urban scenario but lacks considering kinematics. Results are shown in Tab.~\ref{HD}. Note that by incorporating kinematics our method achieves comparatively better performance in all experiments, especially in Exp.4 where the task is more complex. 
	\begin{table}[]
	\centering
	\caption{Trajectory prediction performance comparison on test set using Hausdorff Distance (HD).}
\begin{tabular}{|c|c|c|}
\hline
Experiment     & Wulfmeier et al. \cite{7759328}& Ours            \\ \hline
Exp.1 (E1) & 6.8803                  & 4.2696 \\ \hline
Exp.2 (E2) & 7.2213                  & 4.2000 \\ \hline
Exp.3 (E3) & 5.1404                  & 4.6698 \\ \hline
Exp.4 (E4) & 9.3647                  & 2.2606 \\ \hline
\end{tabular}
\label {HD}
\end{table}

\subsection{Computation Efficiency Analysis}
%We make quantitive comparison of test computation time. 
We choose average time of two stages(i.e. RL and Svf) spent on each sample as comparison metric. %In training both RL and Svf stages are needed while only RL stage is needed for testing. 
In our proposed method, Algorithm 3 and Algorithm 4 are used for RL stage and Svf stage respectively, while Deep IRL uses Algorithm 1 and Algorithm 2. Note that calculating optimal policy and replanning happens at every input sample. The experiment is conducted on an Intel Xeon E5 CPU and an NVIDIA TiTanX GPU. Results are shown in Tab.~\ref{time_train}. Compared to Deep IRL, our method using only CPU takes longer time and this time can be furthermore decreased by using GPU. However, please note that Deep IRL	 doesn't consider kinematics at all and has a much smaller state-action space than ours. We consider kinematics and meanwhile achieve relatively efficient computation by utilizing convolution.
	\begin{table}[h]
	\centering
	\caption{Time spent on two computation stages of each sample}
	\begin{tabular}{|c|c|c|c|}
		\hline
		& \multicolumn{3}{c|}{Train Time}              \\ \hline
		Stage & IRL(without kinematics) \cite{7759328} & Ours(CPU)& Ours(GPU)\\ \hline
		RL  & 0.2647s        & 0.7501s   & 0.3958s   \\ \hline
		Svf & 0.1635s        & 1.0671s   & 0.5161s   \\ \hline
	\end{tabular}
	\label{time_train}
\end{table}
\vspace{-3mm}
%\begin{table}[h]
%	\centering
%	\caption{Test time spent on two computation stages of each sample}
%	\begin{tabular}{|c|c|c|c|}
%		\hline
%		& \multicolumn{3}{c|}{Test Time}               \\ \hline
%		Stage & loop-based Deep IRL & Ours(CPU) & Ours(GPU) \\ \hline
%		RL  & 77.5618s        & 0.6027s   & 0.2541s   \\ \hline
%		Svf & 18.7573s        & 0.8479s   & 0.3860s   \\ \hline
%	\end{tabular}
%	\label{time_test}
%\end{table}

\section{Conclusion and Future works}
\label{sec:conclusion and future works}
A method of off-road traversability analysis and trajectory planning using Deep Maximum Entropy Inverse Reinforcement Learning is proposed. A major novelty is the efficient incorporating of vehicle kinematics. Two convolutional neural networks, i.e., RL ConvNet and Svf ConvNet, are developed that encode vehicle kinematics into convolution kernels, so as to solve the exponential increase of state-space complexity problem and achieve efficient computation in forward reinforcement learning. 
%Experiments are conducted, where four traverse cost functions are learned from demonstration trajectories that represent the behaviors of normal driving, avoid negative obstacles, cross all negative obstacles and cross the negative obstacles if they block the way.
 Results of conducted experiments demonstrate that the learned cost functions are able to guide trajectory planning in compliance with the expert's behaviors and the method has scalability at various scenes. The proposed method achieves improvements on trajectory prediction performance, meanwhile the computation time does not significantly increase. In future work, more extensive experimental studies will be conducted, and improvement on the accuracy of kinematic kernel will be addressed.

\bibliographystyle{IEEEtran}
\bibliography{ref}
%\section{Appendix}
\end{document}